%% file: main.tex
\pdfoutput=1

\documentclass[11pt]{article}

\usepackage[final]{acl}

\usepackage{times}
\usepackage{latexsym}

\usepackage[T1]{fontenc}

\usepackage[utf8]{inputenc}

\usepackage{microtype}

\usepackage{inconsolata}

\usepackage{graphicx}
\usepackage{amsmath}
\usepackage{tabularray}
\usepackage{longtable}
\usepackage{xcolor}
\usepackage{colortbl}
\usepackage{multicol}
\usepackage{booktabs}
\usepackage{makecell}
\usepackage{amsmath, amssymb}
\usepackage{multirow}
\usepackage{mathtools}
\usepackage{enumitem}
\usepackage{subcaption}
\usepackage{float}
\usepackage{graphicx}
\usepackage{caption} 
\usepackage{upgreek}
\usepackage{seqsplit}
\usepackage{color,soul}
\usepackage{arydshln}

\title{LG AI Research \& KAIST at EHRSQL 2024: \\ Self-Training Large Language Models with Pseudo-Labeled Unanswerable Questions for a Reliable Text-to-SQL System on EHRs}

\author{
 \textbf{Yongrae Jo\textsuperscript{1}\thanks{These authors contributed equally to this work.}} \quad
 \textbf{Seongyun Lee\textsuperscript{2}\footnotemark[1]} \quad
 \textbf{Minju Seo\textsuperscript{2}\footnotemark[1]} \quad
 \textbf{Sung Ju Hwang\textsuperscript{2}} \quad
 \textbf{Moontae Lee\textsuperscript{1}}
\\
\\
 \textsuperscript{1}LG AI Research,
 \textsuperscript{2}KAIST
\\
\\
 \texttt{\{yongrae.jo, moontae.lee\}@lgresearch.ai \quad sjhwang82@kaist.ac.kr}
}

\newcommand{\Ours}{\textsc{PLUQ}}
\begin{document}
\maketitle
\begin{abstract}
Text-to-SQL models are pivotal for making Electronic Health Records (EHRs) accessible to healthcare professionals without SQL knowledge. With the advancements in large language models, these systems have become more adept at translating complex questions into SQL queries. Nonetheless, the critical need for reliability in healthcare necessitates these models to accurately identify unanswerable questions or uncertain predictions, preventing misinformation. To address this problem, we present a self-training strategy using pseudo-labeled unanswerable questions to enhance the reliability of text-to-SQL models for EHRs. This approach includes a two-stage training process followed by a filtering method based on the token entropy and query execution. Our methodology's effectiveness is validated by our top performance in the EHRSQL 2024 shared task, showcasing the potential to improve healthcare decision-making through more reliable text-to-SQL systems.

\end{abstract}

\section{Introduction} 

Electronic Health Records (EHRs) are relational databases storing patients' medical histories within hospitals, covering details from admission to discharge. Common challenges with EHRs include difficulties in documenting and tracking health information, supporting team coordination, and sharing data \citep{cifuentes2015electronic}. Although ensuring the accurate capture of relevant information is crucial for addressing these challenges, accessing and querying these records often requires knowledge of SQL, making it challenging for healthcare provides in practical settings without technical expertise. A solution to this problem is developing a text-to-SQL model that can translate natural language questions into SQL queries to retrieve information from EHRs.


Recent advancements in Large Language Models (LLMs) have expanded their utility beyond natural language processing to include code generation, enabling them to interpret text for table manipulation and translate descriptions into code effectively \citep{lee2024learning}. These capabilities showcased by code-generating LLMs \citep{li2023starcoder, roziere2023code, guo2024deepseek} demonstrate their potential in text-to-SQL applications. These developments suggest a promising horizon for leveraging LLMs to make EHR data more accessible to healthcare professionals, eliminating the prerequisite of SQL knowledge and significantly simplifying information retrieval \citep{hwang2019comprehensive, hydranet, mimicsql, kgbaqa-ehr}.

However, in the healthcare domain, the reliability of text-to-SQL models is crucial compared to other areas of NLP application. These models must not only generate precise SQL queries from natural language but also identify unanswerable questions—queries that cannot be solved with the available database—to avoid potentially harmful outcomes. The risk of providing answers to such questions underscores the need for these models to err on the side of caution, by preferring not to provide an answer rather than risking the provision of incorrect information \citep{ehrsql}. This approach underlines the unique challenges faced in healthcare NLP, emphasizing the need for accuracy and the ability to recognize when it cannot provide a reliable answer.


Our work introduces {\Ours}, an approach leveraging the self-training paradigm to improve the reliability of text-to-SQL models for EHRs through training with \textbf{P}seudo-\textbf{L}abeled \textbf{U}nanswerable \textbf{Q}uestions. Self-training, a semi-supervised learning technique, involves re-training a model using its own predictions on unlabeled data to boost its performance. Our method adopts a two-stage version of self-training process, where we initially fine-tune a seed model using a given training dataset. We then augment the training dataset by incorporating unanswerable questions that the fine-tuned model identifies from an unlabeled dataset. Subsequently, we fine-tune the model once more using this augmented training data to produce the final model.

Self-training is commonly used in scenarios where unlabeled data is abundant but obtaining labeled data is costly. Text-to-SQL for EHRs exemplifies such a scenario. In this context, a real-world service can collect users' natural language queries without much effort, but determining the correct SQL statement or verifying its answerability with the given database is time-consuming. We adopt the self-training approach to effectively address the issue of class imbalance between answerable and unanswerable questions, thus enhancing the robustness and reliability of the model's performance.

After the two-stage self-training process, we apply a filtering strategy to eliminate uncertain predictions. This strategy employs two types of filtering: one based on the maximum entropy of tokens and the other on the execution results of queries. Specifically, we assess the entropy in each token generated by the language model, designating the prediction's entropy as that of the token with the highest entropy. If a prediction's entropy exceeds a certain threshold, we consider it an unanswerable question, reflecting the model's lack of confidence in providing a correct answer. Additionally, we remove SQL queries that either produce errors or fail to retrieve valid values from the MIMIC-IV\footnote{https://physionet.org/content/mimic-iv-demo/2.2/} dataset.

This approach was validated by our performance in the EHRSQL 2024 shared task \citep{lee2024overview}, where we achieved the top ranking, demonstrating the effectiveness of our method in improving the reliability of text-to-SQL systems in healthcare.

In summary, our study contributes a method that enhances the reliability of text-to-SQL systems for EHRs, addressing the shared task of handling unanswerable questions. This work supports better access to and utilization of EHRs, aiding in informed healthcare decision-making.

The main contributions of our paper are:

\begin{description}
\item 1. We propose a self-training method that uses pseudo-labeled unanswerable questions to train text-to-SQL models. This approach helps improve the model's ability to identify queries it cannot answer accurately, thereby increasing reliability.
\item 2. We detail the comprehensive strategy employed, from the initial prompting of the model to the filtering steps, to ensure the research can be reproduced. This clarity in methodology allows for the approach to be validated and applied by others in the field, enhancing text-to-SQL systems in healthcare.
\item 3. Our method won the EHRSQL 2024 shared task, demonstrating its practical effectiveness in a competitive setting. This success showcases its potential to contribute to the healthcare field by improving access to EHRS through reliable text-to-SQL systems.
\end{description}

\input{Figures/method_fig}

\section{Related Work} 

In the field of Natural Language Processing (NLP), recent research has focused on text-to-SQL and applying large language models (LLMs) to Electronic Health Records (EHRs). These studies have advanced the handling of complex queries and the processing of healthcare data, setting the stage for our research on test-time data sample labeling and augmentation in EHRs.

\paragraph{Text-to-SQL}
In the evolving field of natural language processing, the development of text-to-SQL technologies represents a significant advancement. Pioneering efforts in this area, \citet{SQLova} and \citet{hydranet}, harnessed the power of BERT for column classification to tackle the Wiki SQL\cite{wikisql} dataset, which is characterized by its simplistic select/where queries. For more complex scenarios, the SPIDER dataset\cite{spider}, comprising Multi-Table questions, necessitated a understanding of relationship between different tables. \citet{ratsql} employed graph-based methods to integrating information, while \citet{bridge} introduced schema linking as an input. Moreover, fine-tuning pre-trained language models such as T5 \cite{t5} has yielded substantial performance improvements in this field.

\paragraph{Large Language Models in Text-to-SQL}
The emergence of LLMs has inspired novel approaches to text-to-SQL tasks. \citet{c3} introduced efficient zero-shot framework, which capitalize on the robust understanding capabilities of LLMs, with a particular emphasis on prompt-based techniques that demonstrates remarkable efficiency. \citet{prompt1, prompt2, prompt3} explore optimal demonstrations based on methodologies like text dense similarity or query similarity selection. \citet{dinsql} enhances the integrity of generated SQL through decomposition of queries and self-correction strategies. \citet{ehragent} proposed LLMs as an agent for generating code and executing it, leverage the few-shot learning capabilities of LLMs for solving the multi-tabular health record datasets.

\paragraph{Enhancement of LLMs Through Data Augmentation and Self-Training}
\citet{self-training} presents an extensive review of self-training methods, including consistency-based approaches and transductive learning. Post the rise of LLMs, the field of self-training methods has garnered considerable attention. To enhance the capabilities of LLMs, some studies have focused on autonomous data generation. \citet{self-instruct} stands out by generating synthetic data from a pool consisting not only of seed data but also data generated through an iterative process. \citet{rada} employs a few-shot learning approach, drawing samples from external sources to align the seed data in low-resource settings. The line of autonomously augmenting data broadens and enhances the model's capabilities. \citet{self-rewarding} introduces using their own outputs to continuously improve both their instruction-following and reward-modeling abilities, demonstrating significant performance enhancements over traditional training methods.

\paragraph{NLP in EHRs}
The application of NLP techniques in EHRs has been extensively explored, utilizing texts and structured knowledge. \citet{emrqa} proposed a question-answering system based on unstructured clinical notes. 
More recently, many works have been developed in the development of generation task based on structured EHRs. \citet{mimicsql} construct the table-based QA datasets using MIMIC-III\cite{mimic-iii} . \citet{kgbaqa-ehr} introduced a graph-based EHR QA system that leverages SPARQL queries from the MIMIC-SQL dataset\cite{mimicsql}. \citet{emrkbqa} focuses on QA tasks using the structured patient records in the MIMIC-III. \citet{ehrsql} datasets containing multi-table queries and those involving null values, reflecting real-world scenarios in healthcare domain.

In our research, we utilize a trained LLM on the training dataset for test-time data sample labeling and subsequent augmentation. This approach is particularly focused on addressing the imbalance in the `null' class.

\section{Method} 
\label{sec:method}
We train a seed model using the original training dataset and then use this model for pseudo labeling on the test set. From this, we select only the samples labeled as unanswerable and augment them to the original training dataset to create the final dataset for self-training. Our self-trained model, {\Ours} generates SQL queries. We apply post-processing and two stages of filtering to these queries to ensure their reliability and produce the final answers.
\subsection{Seed model fine-tuning}
\label{sec:seed_model_ft}
In developing a model specialized for the text-to-SQL task, we initially fine-tuned seed model on the given training data. Because it is widely recognized that there exists a performance gap between open-source LLMs and proprietary LLMs in many benchmarks. In section~\ref{sec:ablation}, the results regarding performance corresponding to changes in the model substantiate it. Therefore, we utilized the Finetuning API provided by OpenAI to fine-tune the GPT-3.5-Turbo-0125 model.

The training dataset comprises a total of 5,124 samples, including both answerable and unanswerable questions. We employed all of these data samples in our training. Furthermore, to ensure that {\Ours} accurately references the correct column names when generating SQL queries, we converted the table schema of the provided MIMIC-IV demo database into text format and incorporated it into the input for training. Additionally, to enhance {\Ours} capability in distinguishing between answerable and unanswerable questions, we incorporated information about unanswerable questions into the input. This strategic inclusion aimed at refining {\Ours} discernment, thus improving its overall accuracy in classifying questions. 
\subsection{Self-training}
\label{sec:self_training}
\paragraph{Unanswerable Question Pseudo Labeling} Unanswerable questions refer to queries that either do not align with the given table schema or require external knowledge, rendering them unsolvable using only the MIMIC-IV demo database for SQL query generation. In our training dataset, the number of answerable questions is considerable, reaching 5,124, whereas unanswerable questions are limited to just 450. This disparity highlights a data imbalance issue within our training dataset, which may impede the model's ability to correctly respond to unanswerable questions during testing.

Moreover, there is a low similarity between the queries in the training data and those in the development/test sets. We found that the average cosine similarity between query embeddings in the train and development sets is only 0.36, and between the train and test sets is 0.34, measured using OpenAI's text-embedding-3-large embedding model. Such a disparity in dataset distribution could lead to significant performance declines for the model at test time. To address these issues, we initially perform pseudo labeling on the development/test set using {\Ours}, which was originally trained solely on the original training dataset.
\paragraph{Training With Augmented Data}
\label{sec:training_w_aug_data}
Pseudo-labeling is one of the techniques used in semi-supervised learning, serving as a powerful tool for addressing issues of data scarcity and label imbalance. Particularly with the EHRSQL dataset, a notable disparity exists: the quantity of unanswerable questions is significantly lower compared to answerable ones within the training data. Training a model with such data increases the likelihood of the model's inability to accurately respond to unanswerable questions. In tasks where reliability is crucial, especially compared to other domains, this could result in substantial penalties. Therefore, we choose to augment the original training set with those samples predicted as unanswerable. Finally, we fine-tune {\Ours} using the augmented dataset.
\subsection{Filtering}
\label{sec:filtering}
Despite the two-stage training process, including self-training, {\Ours} still generates incorrect SQL queries. To enhance the reliability of our final predictions, we implemented a filtering process to sift out samples that were either inaccurately generated or produced with uncertainty by the model. This filtering stage plays a crucial role in ensuring the outputs of {\Ours} are more dependable and accurate.
\paragraph{Maximum Token Entropy Based Filtering}
\label{sec:max_token_ent}
Tokens in a language model-generated output have higher entropy when the information is uncertain. Therefore, treating samples with high entropy as unanswerable questions aids in creating a more reliable system while incurring fewer penalties. We evaluate the entropy of each token produced by the language model, and define the entropy of the prediction based on the token exhibiting the maximum entropy. Then, in the entire set of predictions, samples exceeding a certain entropy level are considered as unanswerable questions and are filtered out. We have set a threshold for this filtering process, determined by the proportion of unanswerable questions in the dataset we aim to predict. This proportion of unanswerable questions is used as a hyperparameter to calibrate the threshold for filtering.
\paragraph{Execution Based Filtering}
\label{sec:execution}
Finally, we implement an additional process of filtering to ensure that the remaining SQL queries, after the initial filtering, can successfully access the MIMIC-IV demo database and retrieve valid values. Utilizing the sqlite3 library in Python, we test each SQL query. Queries that trigger errors, return empty values, or yield None are deemed unable to retrieve valid values. Consequently, we filter these queries as unanswerable questions. This step further ensures the accuracy and reliability of the system by only allowing queries that can effectively interact with the database.

\section{Experiments} 
\label{sec:experiments}
The experiments are conducted on the development and test sets provided by the EHRSQL 2024 shared tasks. All results presented are derived from runs on the official platform. Section~\ref{sec:settings} details the models, datasets, and metrics used for training and inference, while section~\ref{sec:results} discusses the experimental results. 
Finally, in Section~\ref{sec:ablation}, we conduct ablation studies on various components of {\Ours} to examine their individual contributions and impacts.
\input{Tables/main}
\input{Tables/model_ablation}
\input{Tables/prompt_ablation}
\input{Tables/filtering_ablation}
\subsection{Settings}
\label{sec:settings}
\paragraph{Dataset \& Model} We utilize the EHRSQL 2024 dataset for both training and evaluation. The dataset comprises 5,124 training, 1,163 development, and 1,167 test data entries. Notably, only the training dataset is accompanied by gold SQL queries and their corresponding executed gold answers. For questions deemed answerable, it's essential to generate the correct SQL query. For those classified as unanswerable, a null output is required. Database for SQL query generation is the MIMIC-IV demo database. We employ the GPT-3.5-Turbo-0125 model for fine-tuning purposes. Evaluation of our method is conducted on the codabench platform, where we submitted SQL queries predicted by {\Ours} for the test set and obtained scores based on their performance. 
\paragraph{Metrics} 
\begin{figure}[ht]
\centering
\[
\phi_c (x) = 
\begin{cases}
1 & \text{if } x \in \mathcal{Q}_{\text{ans}};g(x) = 1; \text{Acc}(x) = 1\\
0 & \text{if } x \in \mathcal{Q}_{\text{ans}};g(x) = 0,\\
-c & \text{if } x \in \mathcal{Q}_{\text{ans}}; g(x) = 1; \text{Acc}(x) = 0\\
-c & \text{if } x \in \mathcal{Q}_{\text{una}};g(x) = 1,\\
1 & \text{if } x \in \mathcal{Q}_{\text{una}};g(x) = 0.
\end{cases}
\]
\caption{\textbf{Formal Definition of RS} for a single data instance. \( Q_{\text{una}} \) denotes unanswerable question, \( Q_{\text{ans}} \) represents answerable question. \(g(x)=1\) means that model generates SQL query and \(g(x)=0\) denotes that model generates `null'. \(Acc(x) = 1\) signifies instances where the model's prediction is correct, while \(Acc(x) = 0\) indicates cases where the prediction is incorrect. \(c\) represents the penalty.}
\label{fig:rs}
\end{figure}

We utilize the Reliability Score (RS) as our primary metric \citep{ehrsql}. In figure~\ref{fig:rs}, the RS aims to accomplish two main objectives: firstly, it provides rewards for correctly generating SQL for answerable questions \( Q_{\text{ans}} \) and for not generating SQL for unanswerable questions  \( Q_{\text{una}} \); secondly, it imposes penalties for wrongly generating SQL for \( Q_{\text{ans}} \) and for any attempts to create SQL for \( Q_{\text{una}} \). However, the RS neither rewards nor penalizes for choosing not to answer \( Q_{\text{ans}} \). The penalties are structured as 0, 5, 10, or N, where N corresponds to the total number of entries in the dataset. The final score is calculated by adding 1 point for each correct sample and deducting points based on the penalty for incorrect ones, followed by averaging these scores. Importantly, in the EHRSQL 2024 shared task, the primary metric for determining rankings is RS(10).

\begin{figure*}[t!]
    \centering
    \includegraphics[width=0.975\linewidth]{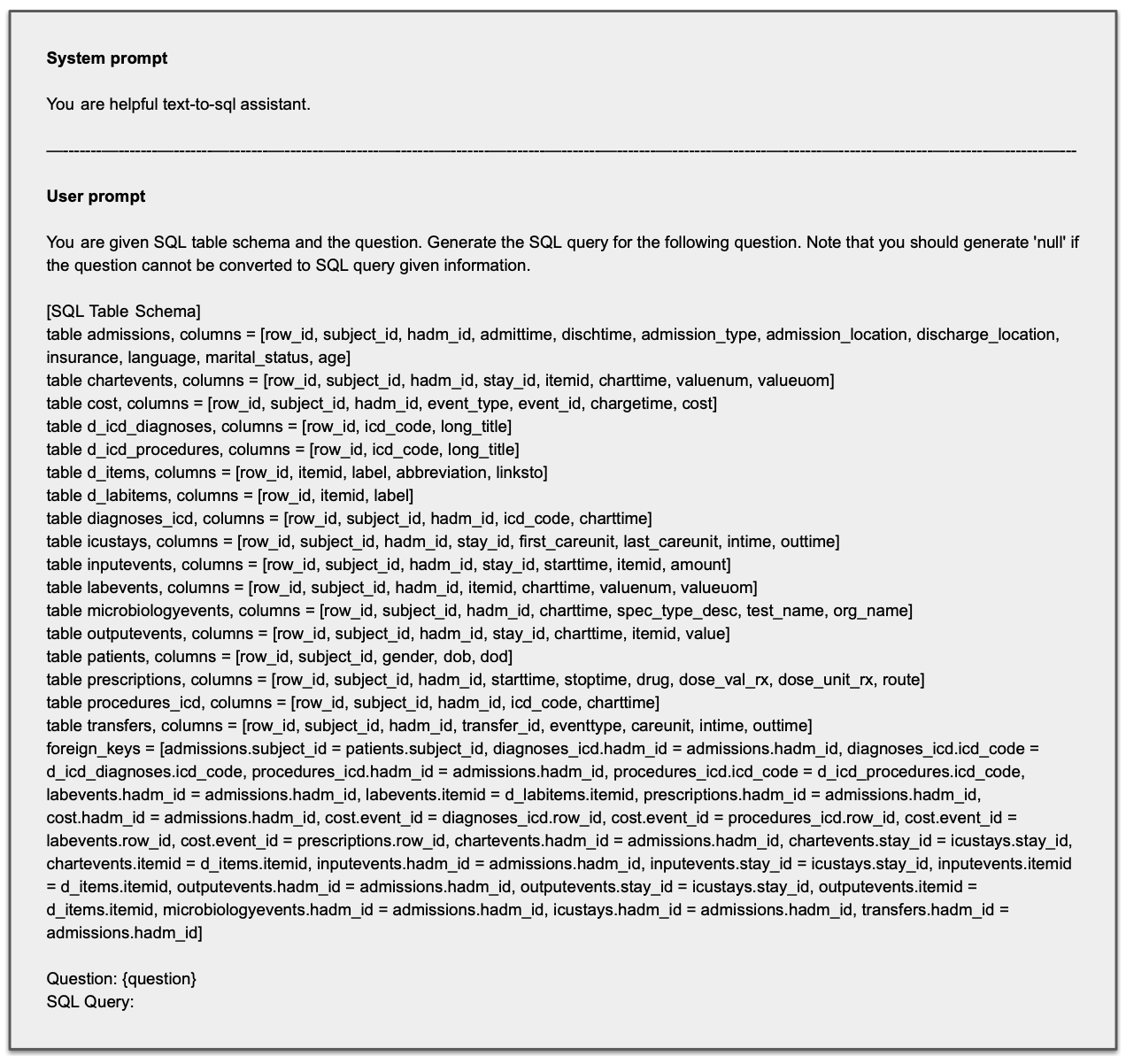}
    \vspace{-0.125in}
    \caption{The system prompt and the user prompt template used in {\Ours}. The prompt integrates instructions for handling unanswerable questions and the MIMIC-IV database schema.}
    \label{fig:prompt}
    \vspace{-0.20in}
\end{figure*}

\subsection{Results}
\label{sec:results}
\paragraph{Development Set} In the development set, {\Ours} exhibits the highest performance in RS(10), the primary metric, which positions it at the top of the official leaderboard when compared with other models. A notable aspect of {\Ours} is the minimal difference between its RS(0) and RS(10) scores compared to other models. This indicates that {\Ours} effectively reduces penalties by categorizing uncertain outcomes in answerable questions and unanswerable questions as 'unanswerable.' This strategy underscores our model's superior reliability, as it avoids the risk of incorrect answers where uncertainty exists, a feature that sets it apart from its counterparts.
\paragraph{Test Set} In the final ranking phase of the shared task, which utilized the test set, {\Ours} experienced a slight overall decrease in scores compared to its performance in the development set. Despite this dip, it maintained a higher score across all RS, including the pivotal RS(10), when compared with other models. This consistent performance across all metrics, even amidst a minor decline, ultimately led {\Ours} to win the EHRSQL 2024 shared task.
\subsection{Ablation Studies}
\label{sec:ablation}
\paragraph{Model Ablation} We observe the performances across difference models. A total of three models were used, namely Flan-T5-base, Tulu-7b, GPT-3.5-Turbo-0125, and GPT-4-Turbo-Preview. Flan-T5-base, Tulu-7b, and GPT-3.5-Turbo-0125 is fine-tuned, while GPT-4-Turbo-Preview is applied with in-context learning. All results are conducted on the development set.

Among the fine-tuned models, GPT-3.5-Turbo-0125 demonstrates the highest performance. This indicates that there is still a performance gap between proprietary and open-source models. Furthermore, despite having more parameters, Tulu-7b shows lower performance compared to Flan-T5-base. Additionally, it is observed that GPT-4-Turbo-Preview, known for its high performance in numerous benchmarks, scored lower than fine-tuned models when only in-context learning is applied.
\paragraph{Prompt Ablation} In the study, we compare the performance of models based on the information included in the input prompts during training. When table schema information is incorporated into the prompts, the models perform better than without it. This suggests that providing table schema information, such as column names, offers a valuable learning signal to the models.

Additionally, explicitly including information about unanswerable questions results in higher scores than when such information is omitted. By providing criteria for answerable and unanswerable questions, the models are aided in avoiding questions they could not answer and focusing on providing accurate responses to those that are answerable.

In the final version of the prompt, we incorporated the database schema of MIMIC-IV as well as the instruction related to unanswerable questions. You can find the prompt in Figure~\ref{fig:prompt}.
\paragraph{Filtering Ablation} 
In table~\ref{tab:filtering_ablation}, by applying execution filtering, which treats invalid SQL queries that either do not execute or retrieve empty values as unanswerable questions, a significant performance improvement is observed, particularly in scenarios with substantial penalties such as RS(10) and RS(N). Additionally, by implementing entropy-based filtering, which filters out SQL queries with higher entropy than a set threshold among those with high maximum token entropy, performance is further enhanced by effectively eliminating SQL queries that, even when executed, return incorrect values.

\section{Conclusion}
\label{sec:conclusion}

In our work, we develop a self-training strategy designed to enhance the reliability of text-to-SQL models for Electronic Health Records (EHRs) through the inclusion of pseudo-labeled unanswerable questions. This approach is particularly valuable in scenarios where there is an abundance of unlabeled data and labeling is costly, thus providing substantial clinical utility in real-world applications. Our approach employs a two-stage training process alongside a filtering mechanism based on token entropy and query execution outcomes to improve the model’s precision and its ability to identify unanswerable questions. The performance is validated by our leading performance in the EHRSQL 2024 shared task. Our method contributes towards rendering EHRs more accessible to healthcare professionals without SQL knowledge, addressing a critical need for reliable information retrieval in healthcare. Future research could explore how large language models facilitate the integration of unstructured medical texts into specific schemas, enhancing interoperability in varied healthcare settings.

\section*{Limitations}
\label{sec:limitation}

Our method achieve the best score in this challenge, as we adopt various techniques to enhance reliability. However, there are some limitations to our approach. Since our model is fine-tuned using EHRs, its ability to generalize across the entire EHR dataset is limited. Additionally, the fine-tuning process requires training data, which poses a challenge due to the high costs and time associated with data collection. Furthermore, despite achieving the highest score among all teams, our RS(N) score still remains negative, indicating that caution should be exercised when considering the application of our method in real-world scenarios.

\section*{Acknowledgements}
We would like to express our gratitude to Professor Minjoon Seo for his invaluable contributions to this project. His guidance and insightful discussions significantly enhanced our research.

\bibliography{custom}



\end{document}

%% file: Figures/method_fig.tex
\begin{figure*}[t!]
    \centering
    \includegraphics[width=0.975\linewidth]{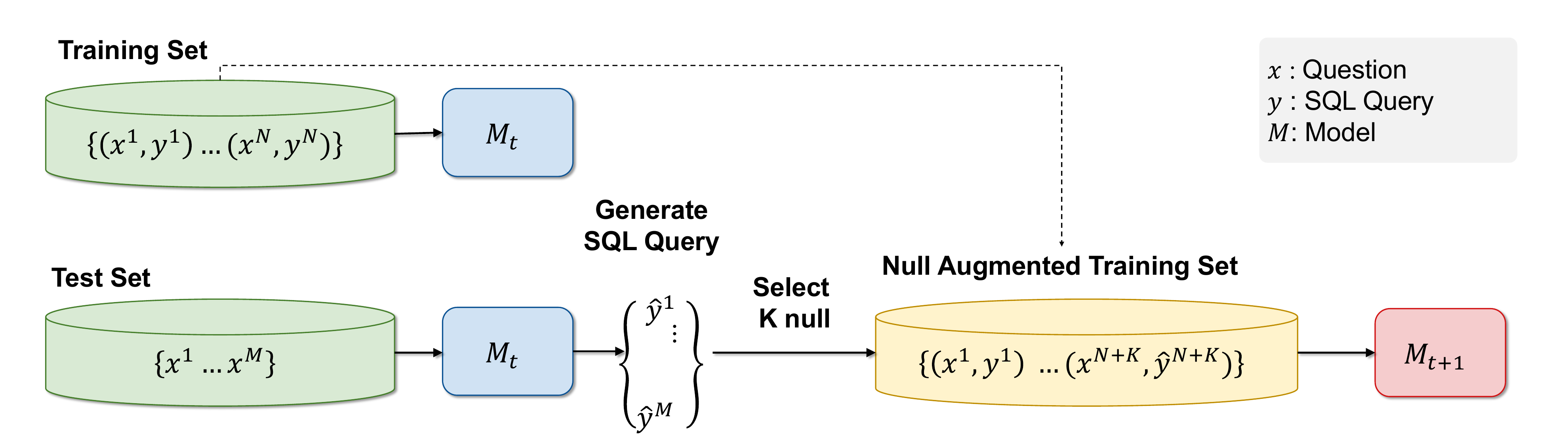}
    \vspace{-0.125in}
    \caption{Training Process and SQL Query Generation. The model is initially trained using the training set. Then, a SQL query (or null) is generated for each sample in the test set using the trained model. Subsequently, we select $K$ null samples and add them to the training set, resulting in a null-augmented training set. This augmented dataset is then used to train the final model, denoted as $M_{t+1}$.}
    \label{fig:method}
    \vspace{-0.20in}
\end{figure*}

%% file: Tables/main.tex
\begin{table*}[t!]
\caption{
\textbf{Results of the Development and Test Phases} on the Official Codabench Leaderboard. The best results are highlighted in bold. Pivotal metric is RS(10) in this shared task. Note that Ours score for the development phase differs from the official leaderboard because we didn't add it to the leaderboard.
}
\vspace{-0.1in}
\label{tab:transfer}
\small
\centering
\resizebox{\textwidth}{!}{
\renewcommand{\arraystretch}{1.1}
\renewcommand{\tabcolsep}{3.0mm}
\begin{tabular}{lccccccccc}
\toprule

& \multicolumn{4}{c}{\bf Development} & \multicolumn{4}{c}{\bf Test} \\
\cmidrule(l{2pt}r{2pt}){2-5} \cmidrule(l{2pt}r{2pt}){6-9}
\textbf{Team} & RS(0) & RS(5) & RS(10) & RS(N) & RS(0) & RS(5) & RS(10)& RS(N) \\

\midrule
\midrule

\textbf{{\Ours} (Ours)}
 & 90.37 & \textbf{89.51} & \textbf{88.65} & \textbf{-109.6} & \textbf{88.17} & \textbf{84.75} & \textbf{81.32} & \textbf{-711.83} \\

\noalign{\vskip 0.25ex}\cdashline{1-9}\noalign{\vskip 0.75ex}

PromptMind
 & 66.38 & 59.5 & 52.62 & -1533.62 & 82.6 & 78.75 & 74.89 & -817.4 \\
 	
ProbGate
 & 84.18 & 79.45 & 74.72 & -1015.82 & 81.92 & 78.06 & 74.21 & -818.08 \\ 

KU-DMIS
 & \textbf{91.57} & 82.98 & 74.38 & -1908.43 & 72.07 & 65.64 & 59.21 & -1427.93\\ 

oleg1996
 & 47.03 & 34.14 & 21.24 & -2952.97 & 68.89 & 56.47 & 44.04 & -2831.11 \\ 

LTRC-IIITH
& N/A & N/A & N/A & N/A & 66.84 & 55.27 & 43.7 & -2633.16 \\

Saama Technologies
 & 57.78 & 50.47 & 43.16 & -1642.22 & 53.21 & 44.64 & 36.08 & -1946.79 \\ 

TEAM\_optimist
& N/A & N/A & N/A & N/A & 14.14	& 	-349.61 & -713.37 &  -84885.86 \\ 

\bottomrule

\end{tabular}
}
\vspace{-0.05in}
\end{table*}

%% file: Tables/model_ablation.tex
 
\begin{table}[t!]
\caption{\textbf{Model Ablation Study} of the Development Set across the finetuned open-source  LLMs and in-context learning, finetuned proprietary LLMs. FT denotes the fine-tuning of the model, while ICL represents in-context learning \citep{wei2022chain}. In this work, the number of few-shot examples used for in-context learning is fixed at 4.}
\label{tab:model_ablation}
\small
\centering
\resizebox{0.475\textwidth}{!}{
\renewcommand{\arraystretch}{0.95}
\begin{tabular}{p{2.7cm}cccc}
\toprule
 \textbf{Models} & {\bf RS(0) } &{\bf RS(5) } & {\bf RS(10) } & {\bf RS(N) } \\
 \midrule
 Flan-T5-base FT & 82.11 & 76.53 & 70.94 & -1217.8 \\
 Tulu-7b FT & 10.23 & -38.77 & -87.9 & -11389.7 \\

GPT-4-Turbo-Preview ICL & 63.52 & -118.85 & -301.22 & -186836.4 \\
\noalign{\vskip 0.25ex}\cdashline{1-5}\noalign{\vskip 0.75ex}
GPT-3.5-Turbo-0125 FT (Ours) & \textbf{90.37} & \textbf{89.51} & \textbf{88.65} & \textbf{-109.6} \\

\bottomrule

\end{tabular}
}
\end{table}

%% file: Tables/prompt_ablation.tex
 
\begin{table}[t!]
\caption{\textbf{Prompt Ablation Study} of the Development Set including the integration of table schema and the incorporation of unanswerable information to evaluate the impact of various prompts on model performance. The base model of fine-tuning is GPT-3.5-Turbo-0125 model.}
\label{tab:prompt_ablation}
\small
\centering
\resizebox{0.475\textwidth}{!}{
\renewcommand{\arraystretch}{0.95}
\begin{tabular}{p{2.7cm}cccc}
\toprule
 \textbf{Models} & {\bf RS(0) } &{\bf RS(5) } & {\bf RS(10) } & {\bf RS(N) } \\
 \midrule
Fine-Tuning & 83.23 & 78.5 & 73.77 & -1016.7\\
 + Table Schema & 89.85 & 83.83 & 77.82 & 
 -1310.1 \\
+ Unans Info (Ours) & \textbf{90.37} & \textbf{89.51} & \textbf{88.65} & \textbf{-109.6} \\

\bottomrule

\end{tabular}
}
\vspace{-0.125in}
\end{table}

%% file: Tables/filtering_ablation.tex
 \begin{table}[t!]
\caption{\textbf{Filtering Ablation Study} of Development Set. For maximum token entropy based filtering, we filtered out SQL queries possessing high entropy within the top 7\%, classifying them as unanswerable questions.}
\label{tab:filtering_ablation}
\small
\centering
\resizebox{0.475\textwidth}{!}{
\renewcommand{\arraystretch}{0.95}
\begin{tabular}{p{2.7cm}cccc}
\toprule
 \textbf{Models} & {\bf RS(0) } &{\bf RS(5) } & {\bf RS(10) } & {\bf RS(N) } \\
 \midrule
No Filtering & 80.82 & 5.58 & -69.94 & -17419.1\\
 + Exec Filtering & \textbf{93.98} & \textbf{89.68} & 85.38 & 
 -906.01 \\
+ Ent Filtering (Ours) & 90.37 & 89.51 & \textbf{88.65} & \textbf{-109.6} \\

\bottomrule

\end{tabular}
}
\vspace{-0.125in}
\end{table}

%% file: main.bbl
\begin{thebibliography}{31}
\providecommand{\natexlab}[1]{#1}

\bibitem[{Amini et~al.(2022)Amini, Feofanov, Pauletto, Devijver, and Maximov}]{self-training}
Massih{-}Reza Amini, Vasilii Feofanov, Lo{\"{\i}}c Pauletto, Emilie Devijver, and Yury Maximov. 2022.
\newblock \href {https://arxiv.org/abs/2202.12040} {Self-training: {A} survey}.
\newblock \emph{CoRR}, abs/2202.12040.

\bibitem[{Cifuentes et~al.(2015)Cifuentes, Davis, Fernald, Gunn, Dickinson, and Cohen}]{cifuentes2015electronic}
Maribel Cifuentes, Melinda Davis, Doug Fernald, Rose Gunn, Perry Dickinson, and Deborah~J Cohen. 2015.
\newblock Electronic health record challenges, workarounds, and solutions observed in practices integrating behavioral health and primary care.
\newblock \emph{The Journal of the American Board of Family Medicine}, 28(Supplement 1):S63--S72.

\bibitem[{Dong et~al.(2023)Dong, Zhang, Ge, Mao, Gao, Chen, Lin, and Lou}]{c3}
Xuemei Dong, Chao Zhang, Yuhang Ge, Yuren Mao, Yunjun Gao, Lu~Chen, Jinshu Lin, and Dongfang Lou. 2023.
\newblock \href {https://doi.org/10.48550/ARXIV.2307.07306} {{C3:} zero-shot text-to-sql with chatgpt}.
\newblock \emph{CoRR}, abs/2307.07306.

\bibitem[{Gao et~al.(2024)Gao, Wang, Li, Sun, Qian, Ding, and Zhou}]{prompt3}
Dawei Gao, Haibin Wang, Yaliang Li, Xiuyu Sun, Yichen Qian, Bolin Ding, and Jingren Zhou. 2024.
\newblock \href {https://www.vldb.org/pvldb/vol17/p1132-gao.pdf} {Text-to-sql empowered by large language models: {A} benchmark evaluation}.
\newblock \emph{Proc. {VLDB} Endow.}, 17(5):1132--1145.

\bibitem[{Guo et~al.(2024)Guo, Zhu, Yang, Xie, Dong, Zhang, Chen, Bi, Wu, Li et~al.}]{guo2024deepseek}
Daya Guo, Qihao Zhu, Dejian Yang, Zhenda Xie, Kai Dong, Wentao Zhang, Guanting Chen, Xiao Bi, Y~Wu, YK~Li, et~al. 2024.
\newblock Deepseek-coder: When the large language model meets programming--the rise of code intelligence.
\newblock \emph{arXiv preprint arXiv:2401.14196}.

\bibitem[{Hwang et~al.(2019{\natexlab{a}})Hwang, Yim, Park, and Seo}]{hwang2019comprehensive}
Wonseok Hwang, Jinyeong Yim, Seunghyun Park, and Minjoon Seo. 2019{\natexlab{a}}.
\newblock A comprehensive exploration on wikisql with table-aware word contextualization.
\newblock \emph{arXiv preprint arXiv:1902.01069}.

\bibitem[{Hwang et~al.(2019{\natexlab{b}})Hwang, Yim, Park, and Seo}]{SQLova}
Wonseok Hwang, Jinyeung Yim, Seunghyun Park, and Minjoon Seo. 2019{\natexlab{b}}.
\newblock \href {https://arxiv.org/abs/1902.01069} {A comprehensive exploration on wikisql with table-aware word contextualization}.
\newblock \emph{CoRR}, abs/1902.01069.

\bibitem[{Johnson et~al.(2016)Johnson, Pollard, Shen, Lehman, Feng, Ghassemi, Moody, Szolovits, Anthony~Celi, and Mark}]{mimic-iii}
Alistair~EW Johnson, Tom~J Pollard, Lu~Shen, Li-wei~H Lehman, Mengling Feng, Mohammad Ghassemi, Benjamin Moody, Peter Szolovits, Leo Anthony~Celi, and Roger~G Mark. 2016.
\newblock Mimic-iii, a freely accessible critical care database.
\newblock \emph{Scientific data}, 3(1):1--9.

\bibitem[{Lee et~al.(2023)Lee, Hwang, Bae, Kwon, Shin, Yang, Seo, Kim, and Choi}]{ehrsql}
Gyubok Lee, Hyeonji Hwang, Seongsu Bae, Yeonsu Kwon, Woncheol Shin, Seongjun Yang, Minjoon Seo, Jong{-}Yeup Kim, and Edward Choi. 2023.
\newblock \href {https://doi.org/10.48550/ARXIV.2301.07695} {{EHRSQL:} {A} practical text-to-sql benchmark for electronic health records}.
\newblock \emph{CoRR}, abs/2301.07695.

\bibitem[{Lee et~al.(2024{\natexlab{a}})Lee, Kweon, Bae, and Choi}]{lee2024overview}
Gyubok Lee, Sunjun Kweon, Seongsu Bae, and Edward Choi. 2024{\natexlab{a}}.
\newblock Overview of the ehrsql 2024 shared task on reliable text-to-sql modeling on electronic health records.
\newblock In \emph{Proceedings of the 6th Clinical Natural Language Processing Workshop}, Mexico City, Mexico. Association for Computational Linguistics.

\bibitem[{Lee et~al.(2024{\natexlab{b}})Lee, Kim, Yu, Rossi, and Chen}]{lee2024learning}
Younghun Lee, Sungchul Kim, Tong Yu, Ryan~A Rossi, and Xiang Chen. 2024{\natexlab{b}}.
\newblock Learning to reduce: Optimal representations of structured data in prompting large language models.
\newblock \emph{arXiv preprint arXiv:2402.14195}.

\bibitem[{Li et~al.(2023)Li, Allal, Zi, Muennighoff, Kocetkov, Mou, Marone, Akiki, Li, Chim et~al.}]{li2023starcoder}
Raymond Li, Loubna~Ben Allal, Yangtian Zi, Niklas Muennighoff, Denis Kocetkov, Chenghao Mou, Marc Marone, Christopher Akiki, Jia Li, Jenny Chim, et~al. 2023.
\newblock Starcoder: may the source be with you!
\newblock \emph{arXiv preprint arXiv:2305.06161}.

\bibitem[{Lin et~al.(2020)Lin, Socher, and Xiong}]{bridge}
Xi~Victoria Lin, Richard Socher, and Caiming Xiong. 2020.
\newblock \href {https://doi.org/10.18653/V1/2020.FINDINGS-EMNLP.438} {Bridging textual and tabular data for cross-domain text-to-sql semantic parsing}.
\newblock In \emph{Findings of the Association for Computational Linguistics: {EMNLP} 2020, Online Event, 16-20 November 2020}, volume {EMNLP} 2020 of \emph{Findings of {ACL}}, pages 4870--4888. Association for Computational Linguistics.

\bibitem[{Lyu et~al.(2020)Lyu, Chakrabarti, Hathi, Kundu, Zhang, and Chen}]{hydranet}
Qin Lyu, Kaushik Chakrabarti, Shobhit Hathi, Souvik Kundu, Jianwen Zhang, and Zheng Chen. 2020.
\newblock \href {https://arxiv.org/abs/2008.04759} {Hybrid ranking network for text-to-sql}.
\newblock \emph{CoRR}, abs/2008.04759.

\bibitem[{Nan et~al.(2023)Nan, Zhao, Zou, Ri, Tae, Zhang, Cohan, and Radev}]{prompt2}
Linyong Nan, Yilun Zhao, Weijin Zou, Narutatsu Ri, Jaesung Tae, Ellen Zhang, Arman Cohan, and Dragomir Radev. 2023.
\newblock \href {https://doi.org/10.48550/ARXIV.2305.12586} {Enhancing few-shot text-to-sql capabilities of large language models: {A} study on prompt design strategies}.
\newblock \emph{CoRR}, abs/2305.12586.

\bibitem[{Pampari et~al.(2018)Pampari, Raghavan, Liang, and Peng}]{emrqa}
Anusri Pampari, Preethi Raghavan, Jennifer~J. Liang, and Jian Peng. 2018.
\newblock \href {https://doi.org/10.18653/V1/D18-1258} {emrqa: {A} large corpus for question answering on electronic medical records}.
\newblock In \emph{Proceedings of the 2018 Conference on Empirical Methods in Natural Language Processing, Brussels, Belgium, October 31 - November 4, 2018}, pages 2357--2368. Association for Computational Linguistics.

\bibitem[{Park et~al.(2021)Park, Cho, Lee, Choo, and Choi}]{kgbaqa-ehr}
Junwoo Park, Youngwoo Cho, Haneol Lee, Jaegul Choo, and Edward Choi. 2021.
\newblock \href {https://proceedings.mlr.press/v149/park21a.html} {Knowledge graph-based question answering with electronic health records}.
\newblock In \emph{Proceedings of the Machine Learning for Healthcare Conference, {MLHC} 2021, 6-7 August 2021, Virtual Event}, volume 149 of \emph{Proceedings of Machine Learning Research}, pages 36--53. {PMLR}.

\bibitem[{Pourreza and Rafiei(2023)}]{dinsql}
Mohammadreza Pourreza and Davood Rafiei. 2023.
\newblock \href {http://papers.nips.cc/paper\_files/paper/2023/hash/72223cc66f63ca1aa59edaec1b3670e6-Abstract-Conference.html} {{DIN-SQL:} decomposed in-context learning of text-to-sql with self-correction}.
\newblock In \emph{Advances in Neural Information Processing Systems 36: Annual Conference on Neural Information Processing Systems 2023, NeurIPS 2023, New Orleans, LA, USA, December 10 - 16, 2023}.

\bibitem[{Raffel et~al.(2019)Raffel, Shazeer, Roberts, Lee, Narang, Matena, Zhou, Li, and Liu}]{t5}
Colin Raffel, Noam Shazeer, Adam Roberts, Katherine Lee, Sharan Narang, Michael Matena, Yanqi Zhou, Wei Li, and Peter~J. Liu. 2019.
\newblock \href {https://arxiv.org/abs/1910.10683} {Exploring the limits of transfer learning with a unified text-to-text transformer}.
\newblock \emph{CoRR}, abs/1910.10683.

\bibitem[{Raghavan et~al.(2021)Raghavan, Liang, Mahajan, Chandra, and Szolovits}]{emrkbqa}
Preethi Raghavan, Jennifer~J. Liang, Diwakar Mahajan, Rachita Chandra, and Peter Szolovits. 2021.
\newblock \href {https://doi.org/10.18653/V1/2021.BIONLP-1.7} {emrkbqa: {A} clinical knowledge-base question answering dataset}.
\newblock In \emph{Proceedings of the 20th Workshop on Biomedical Language Processing, BioNLP@NAACL-HLT 2021, Online, June 11, 2021}, pages 64--73. Association for Computational Linguistics.

\bibitem[{Roziere et~al.(2023)Roziere, Gehring, Gloeckle, Sootla, Gat, Tan, Adi, Liu, Remez, Rapin et~al.}]{roziere2023code}
Baptiste Roziere, Jonas Gehring, Fabian Gloeckle, Sten Sootla, Itai Gat, Xiaoqing~Ellen Tan, Yossi Adi, Jingyu Liu, Tal Remez, J{\'e}r{\'e}my Rapin, et~al. 2023.
\newblock Code llama: Open foundation models for code.
\newblock \emph{arXiv preprint arXiv:2308.12950}.

\bibitem[{Seo et~al.(2024)Seo, Baek, Thorne, and Hwang}]{rada}
Minju Seo, Jinheon Baek, James Thorne, and Sung~Ju Hwang. 2024.
\newblock \href {https://doi.org/10.48550/ARXIV.2402.13482} {Retrieval-augmented data augmentation for low-resource domain tasks}.
\newblock \emph{CoRR}, abs/2402.13482.

\bibitem[{Shi et~al.(2024)Shi, Xu, Zhuang, Yu, Zhang, Wu, Zhu, Ho, Yang, and Wang}]{ehragent}
Wenqi Shi, Ran Xu, Yuchen Zhuang, Yue Yu, Jieyu Zhang, Hang Wu, Yuanda Zhu, Joyce~C. Ho, Carl Yang, and May~D. Wang. 2024.
\newblock \href {https://doi.org/10.48550/ARXIV.2401.07128} {Ehragent: Code empowers large language models for complex tabular reasoning on electronic health records}.
\newblock \emph{CoRR}, abs/2401.07128.

\bibitem[{Tai et~al.(2023)Tai, Chen, Zhang, Deng, and Sun}]{prompt1}
Chang{-}Yu Tai, Ziru Chen, Tianshu Zhang, Xiang Deng, and Huan Sun. 2023.
\newblock \href {https://aclanthology.org/2023.emnlp-main.327} {Exploring chain of thought style prompting for text-to-sql}.
\newblock In \emph{Proceedings of the 2023 Conference on Empirical Methods in Natural Language Processing, {EMNLP} 2023, Singapore, December 6-10, 2023}, pages 5376--5393. Association for Computational Linguistics.

\bibitem[{Wang et~al.(2020{\natexlab{a}})Wang, Shin, Liu, Polozov, and Richardson}]{ratsql}
Bailin Wang, Richard Shin, Xiaodong Liu, Oleksandr Polozov, and Matthew Richardson. 2020{\natexlab{a}}.
\newblock \href {https://doi.org/10.18653/V1/2020.ACL-MAIN.677} {{RAT-SQL:} relation-aware schema encoding and linking for text-to-sql parsers}.
\newblock In \emph{Proceedings of the 58th Annual Meeting of the Association for Computational Linguistics, {ACL} 2020, Online, July 5-10, 2020}, pages 7567--7578. Association for Computational Linguistics.

\bibitem[{Wang et~al.(2020{\natexlab{b}})Wang, Shi, and Reddy}]{mimicsql}
Ping Wang, Tian Shi, and Chandan~K. Reddy. 2020{\natexlab{b}}.
\newblock \href {https://doi.org/10.1145/3366423.3380120} {Text-to-sql generation for question answering on electronic medical records}.
\newblock In \emph{{WWW} '20: The Web Conference 2020, Taipei, Taiwan, April 20-24, 2020}, pages 350--361. {ACM} / {IW3C2}.

\bibitem[{Wang et~al.(2023)Wang, Kordi, Mishra, Liu, Smith, Khashabi, and Hajishirzi}]{self-instruct}
Yizhong Wang, Yeganeh Kordi, Swaroop Mishra, Alisa Liu, Noah~A. Smith, Daniel Khashabi, and Hannaneh Hajishirzi. 2023.
\newblock \href {https://doi.org/10.18653/V1/2023.ACL-LONG.754} {Self-instruct: Aligning language models with self-generated instructions}.
\newblock In \emph{Proceedings of the 61st Annual Meeting of the Association for Computational Linguistics (Volume 1: Long Papers), {ACL} 2023, Toronto, Canada, July 9-14, 2023}, pages 13484--13508. Association for Computational Linguistics.

\bibitem[{Wei et~al.(2022)Wei, Wang, Schuurmans, Bosma, Xia, Chi, Le, Zhou et~al.}]{wei2022chain}
Jason Wei, Xuezhi Wang, Dale Schuurmans, Maarten Bosma, Fei Xia, Ed~Chi, Quoc~V Le, Denny Zhou, et~al. 2022.
\newblock Chain-of-thought prompting elicits reasoning in large language models.
\newblock \emph{Advances in neural information processing systems}, 35:24824--24837.

\bibitem[{Yu et~al.(2018)Yu, Zhang, Yang, Yasunaga, Wang, Li, Ma, Li, Yao, Roman, Zhang, and Radev}]{spider}
Tao Yu, Rui Zhang, Kai Yang, Michihiro Yasunaga, Dongxu Wang, Zifan Li, James Ma, Irene Li, Qingning Yao, Shanelle Roman, Zilin Zhang, and Dragomir~R. Radev. 2018.
\newblock \href {https://doi.org/10.18653/V1/D18-1425} {Spider: {A} large-scale human-labeled dataset for complex and cross-domain semantic parsing and text-to-sql task}.
\newblock In \emph{Proceedings of the 2018 Conference on Empirical Methods in Natural Language Processing, Brussels, Belgium, October 31 - November 4, 2018}, pages 3911--3921. Association for Computational Linguistics.

\bibitem[{Yuan et~al.(2024)Yuan, Pang, Cho, Sukhbaatar, Xu, and Weston}]{self-rewarding}
Weizhe Yuan, Richard~Yuanzhe Pang, Kyunghyun Cho, Sainbayar Sukhbaatar, Jing Xu, and Jason Weston. 2024.
\newblock \href {https://doi.org/10.48550/ARXIV.2401.10020} {Self-rewarding language models}.
\newblock \emph{CoRR}, abs/2401.10020.

\bibitem[{Zhong et~al.(2017)Zhong, Xiong, and Socher}]{wikisql}
Victor Zhong, Caiming Xiong, and Richard Socher. 2017.
\newblock \href {https://arxiv.org/abs/1709.00103} {Seq2sql: Generating structured queries from natural language using reinforcement learning}.
\newblock \emph{CoRR}, abs/1709.00103.

\end{thebibliography}
